\documentclass{article}
\usepackage{spconf,amsmath,graphicx,hyperref}

\usepackage{cite}
\usepackage{amssymb}
\usepackage{graphicx}
\usepackage{textcomp}
\usepackage{xcolor}

\usepackage{algpseudocode}
\usepackage{algorithm}

\usepackage{booktabs}
\usepackage{multirow}
\usepackage{siunitx}

\usepackage{mathtools}
\usepackage{adjustbox}
\newcommand{\colshrink}[1]{\adjustbox{max width=\linewidth}{$\displaystyle #1$}}
\newcommand{\smat}[1]{\left(\begin{smallmatrix}#1\end{smallmatrix}\right)}

\newenvironment{methodmath}{%
  \begingroup
  \setlength{\abovedisplayskip}{6pt}%
  \setlength{\belowdisplayskip}{6pt}%
  \setlength{\jot}{1pt}%
  \setlength{\arraycolsep}{2pt}%
  \everydisplay{\small}%
}{\endgroup}

\usepackage{booktabs,pifont}
\newcommand{\cmark}{\ding{51}} 
\newcommand{\xmark}{\ding{55}} 

\def\BibTeX{{\rm B\kern-.05em{\sc i\kern-.025em b}\kern-.08em
    T\kern-.1667em\lower.7ex\hbox{E}\kern-.125emX}}

\def\x{{\mathbf x}}

\title{MVP: Motion Vector Propagation for Zero-Shot Video Object Detection}

\name{Binhua Huang\textsuperscript{\dag}, 
        Ni Wang\textsuperscript{\ddag}, 
        Wendong Yao\textsuperscript{\dag}, 
        Soumyabrata Dev\textsuperscript{\dag}
        }
\address{\textsuperscript{\dag}School of Computer Science, University College Dublin \\
        \textsuperscript{\ddag}Amazon Development Center Germany GmbH, Berlin, Germany\\
        }

\begin{document}

\maketitle

\begin{abstract}
Running a large open-vocabulary (Open-vocab) detector on every video frame is accurate but expensive. We introduce a training-free pipeline that invokes OWLv2 only on fixed-interval keyframes and propagates detections to intermediate frames using compressed-domain motion vectors (MV). A simple 3×3 grid aggregation of motion vectors provides translation and uniform-scale updates, augmented with an area-growth check and an optional single-class switch. The method requires no labels, no fine-tuning, and uses the same prompt list for all open-vocabulary methods. On ILSVRC2015-VID (validation dataset), our approach (MVP) attains \(\text{mAP@0.5}=0.609\) and \(\text{mAP@[0.5:0.95]}=0.316\). At loose intersection-over-union (IoU) thresholds it remains close to framewise OWLv2-Large (\(0.747/0.721\) at 0.2/0.3 versus \(0.784/0.780\)), reflecting that coarse localization is largely preserved. Under the same keyframe schedule, MVP outperforms tracker-based propagation (MOSSE, KCF, CSRT) at \(\text{mAP@0.5}\). A supervised reference (YOLOv12x) reaches \(0.631\) at \(\text{mAP@0.5}\) but requires labeled training, whereas our method remains label-free and open-vocabulary. These results indicate that compressed-domain propagation is a practical way to reduce detector invocations while keeping strong zero-shot coverage in videos.  Our code and models are available at \url{https://github.com/microa/MVP}.
\end{abstract}


\section{Introduction}
Recent reports estimate that modern networks transmit enormous volumes of video every minute, placing heavy demands on automated video analytics~\cite{cisco2020cisco}. Among many tasks, video object detection is central to content moderation, intelligent traffic monitoring, and high-level scene understanding. However, adapting conventional object detectors that are designed for static images to the dynamics of video can be prohibitively expensive. Classic video pipelines either require retraining on large, domain-specific datasets~\cite{kang2016object} or depend on temporal modules such as optical flow and sequence-level aggregation~\cite{zhu2017flow,wu2019sequence}, which improve robustness but add significant runtime and often reduce generalization across domains.

In parallel, open-vocabulary detection relaxes the closed-set assumption by transferring semantics from large-scale vision and language pretraining. Representative approaches include CLIP and its detection adaptations such as ViLD, Detic, and RegionCLIP~\cite{radford2021learning,gu2021open,zhou2022detecting,zhong2022regionclip}. OWLv2~\cite{minderer2023scaling} further delivers strong zero-shot detection without task-specific fine-tuning~\cite{minderer2023scaling}. Despite this progress, running a large open-vocabulary detector on every video frame remains impractical in long videos or on edge devices because the per-frame inference cost is high even without additional training.

A complementary line of research reduces computation by exploiting signals that are already present in compressed video. Modern codecs such as H.264 provide motion vectors and residuals that can be repurposed to propagate features or boxes between sparse keyframes, avoiding explicit optical flow and heavy per-frame inference~\cite{wu2018compressed,liu2018mobile,wang2019fast}. Many of these methods, however, assume closed vocabularies or require some training on the target dataset. As a result, a simple training-free recipe that unifies open-vocabulary generality with efficient video inference remains underexplored.

\textbf{This work.} We present a training-free and end-to-end pipeline that combines a pretrained open-vocabulary detector (OWLv2) with a lightweight Motion Vector Propagation (MVP) module. The detector is invoked only on fixed-interval keyframes (every \(K\) frames). For intermediate frames, MVP reads codec motion vectors to propagate and gently refine bounding boxes~\cite{wu2018compressed,wang2019fast}. The design needs no additional labels or fine-tuning, cuts the number of expensive detector calls, and preserves strong coarse detection quality across diverse scenes. In experiments, the fixed keyframe schedule recovers much of the framewise baseline at loose IoUs with an expected drop at tight IoUs due to propagation error, and it outperforms classical tracker-based propagation under the same schedule.

\begin{figure*}[t]
    \centering
    \includegraphics[width=0.95\linewidth]{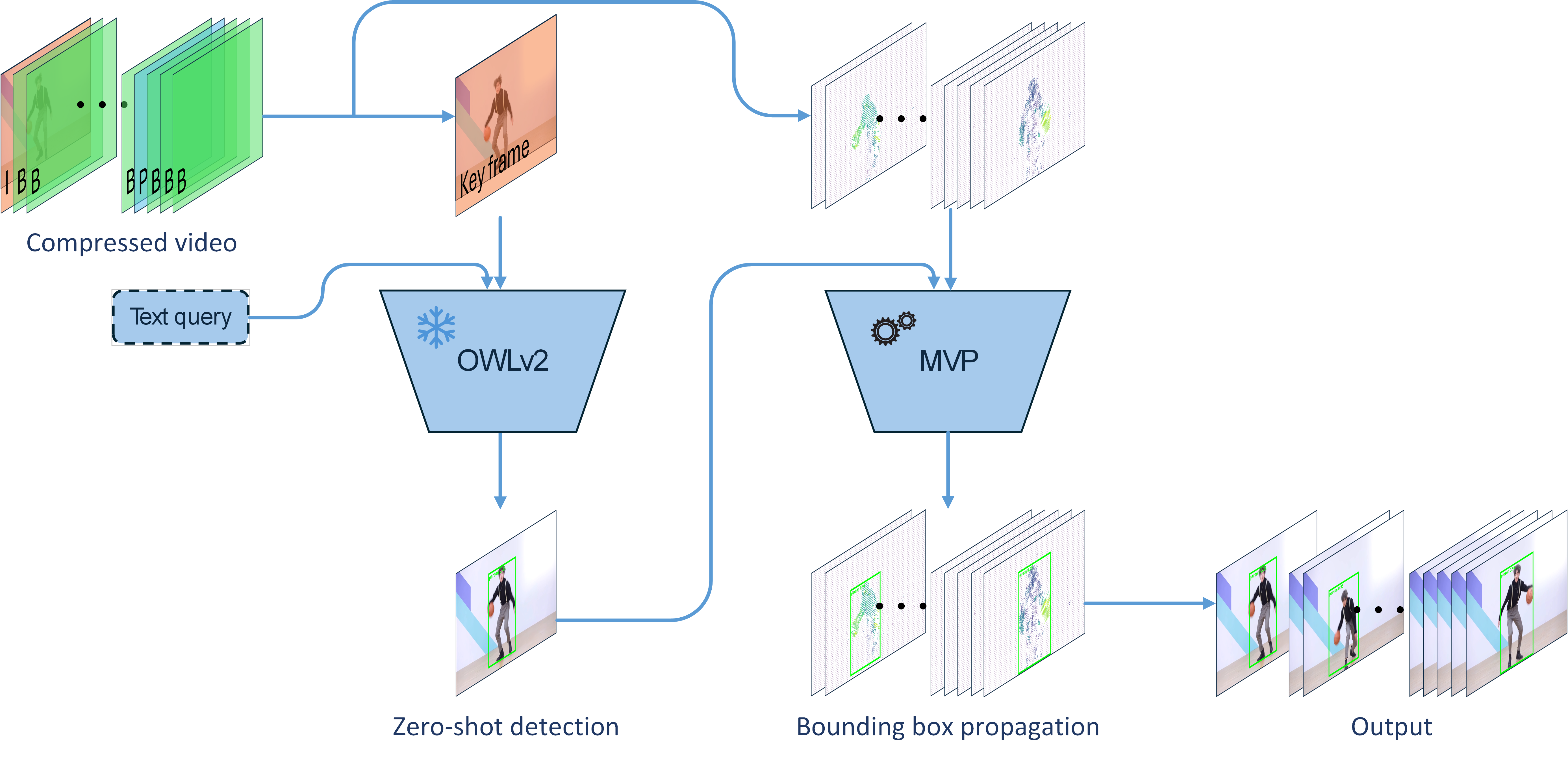}
    \vspace{-5pt}
    \caption{
    \textbf{Overview of the proposed pipeline.}
    Given a compressed video with I-frames and P/B-frames, we select keyframes at a fixed interval (every \(K\) frames) and apply OWLv2~\cite{minderer2023scaling} with a text prompt to obtain zero-shot detections. For intermediate frames, the MVP module uses codec motion vectors to propagate, adjust, and refine bounding boxes~\cite{wu2018compressed,wang2019fast}. This schedule reduces expensive detector calls while maintaining competitive detection quality, enabling a training-free and efficient open-vocabulary video detector.
    }
    \label{fig:main_overview}
    \vspace{-5pt}
\end{figure*}

\textbf{Contributions.}
\begin{itemize}
    \item \textbf{Training-free and plug-and-play.} A video detection pipeline that requires no additional training or fine-tuning and can be deployed on new datasets without labeled data.
    \item \textbf{Fixed-interval keyframe scheduling with codec-driven propagation.} Per-frame detection is replaced by keyframes every \(K\) frames, while compressed-domain motion cues propagate boxes across intermediate frames, reducing computation and retaining competitive accuracy~\cite{wu2018compressed,liu2018mobile,wang2019fast}.
    \item \textbf{Robust propagation techniques.} A practical combination of a 3×3 grid matching strategy, an area-growth check, and a single-class detection mode stabilizes tracks and mitigates drift without supervision.
\end{itemize}



The proposed approach offers a practical path toward scalable, label-free video understanding by unifying open-vocabulary detection~\cite{radford2021learning,gu2021open,zhou2022detecting,zhong2022regionclip,minderer2023scaling} with compressed-domain motion~\cite{wu2018compressed,liu2018mobile,wang2019fast}, matching the growth and diversity of real-world video.

\section{Method}
\begin{methodmath}
We build a training-free zero-shot video detector by combining a pretrained OWLv2 with a lightweight Motion Vector Propagation (MVP) rule. Keyframes are processed by OWLv2, and intermediate frames update boxes using codec motion vectors. This section specifies the representation, the propagation tests, and the fallback and single-class switches in compact mathematical form. All displays use a unified small size, and only overflow formulas shrink to the column width.

\noindent\textbf{Setup and schedule.}
Let frames be \(I_0,\dots,I_{T-1}\) with image size \((W,H)\). Let \(C\) be the set of text categories. We choose a fixed keyframe interval \(K\).
Denote detections at time \(t\) by \(D_t=\{(b^{(t)}_n,s^{(t)}_n,c^{(t)}_n)\}_n\) where each box \(b=(x_c,y_c,w,h)\) follows the YOLO format~\cite{redmon2016you} in $[0,1]^4$, score $s\in[0,1]$, and class \(c\in C\).
Let \(S_t\in\{0,1\}\) denote the single-class switch and \(q_t\) the active prompt set:
\[
\colshrink{
\begin{aligned}
q_t=
\begin{cases}
\{c^\dagger\} & \text{if } S_t=1 \text{ with current class } c^\dagger,\\
C & \text{if } S_t=0.
\end{cases}
\end{aligned}}
\]
\noindent\textbf{Per-box states.} All MVP updates (translation and scale tests, area-growth checks, and fallbacks) are applied independently to each detection \(n\) in \(D_{t-1}\), maintaining per-box states \((t_n^\star, A_n^\star)\). 
The detector is invoked according to
\[
\colshrink{
\begin{aligned}
D_t=
\begin{cases}
\textsc{OWLv2}(I_t,\,q_t) \ \text{if } (t \bmod K = 0)\ \text{or}\ \exists n:\ \mathrm{fallback}_n(t),\\
\mathcal{P}\bigl(D_{t-1},\,\mathrm{MV}(t)\bigr) & \text{otherwise},
\end{cases}
\end{aligned}}
\]
where \(\mathcal{P}\) is the MVP update using motion vectors at time \(t\).

\noindent\textbf{Pixel–YOLO conversions.}
For a pixel box \(b_{\text{pix}}=(x_{\min},y_{\min},\\x_{\max},y_{\max})\), define the mapping to YOLO coordinates and its inverse.
\[
\colshrink{
\Gamma\bigl(b_{\text{pix}};W,H\bigr)=
\smat{
\dfrac{x_{\min}+x_{\max}}{2W}\\[2pt]
\dfrac{y_{\min}+y_{\max}}{2H}\\[2pt]
\dfrac{x_{\max}-x_{\min}}{W}\\[2pt]
\dfrac{y_{\max}-y_{\min}}{H}
}
=
\smat{x_c\\[2pt] y_c\\[2pt] w\\[2pt] h},\qquad 
\Gamma^{-1}\!\smat{x_c\\y_c\\w\\h}:\ 
\left\{
\begin{aligned}
x_{\min}&=W\!\left(x_c-\tfrac{w}{2}\right),&
x_{\max}&=W\!\left(x_c+\tfrac{w}{2}\right),\\
y_{\min}&=H\!\left(y_c-\tfrac{h}{2}\right),&
y_{\max}&=H\!\left(y_c+\tfrac{h}{2}\right).
\end{aligned}
\right.
}
\]

\noindent\textbf{3×3 MV Aggregation.}
Given a previous box \(b^{(t-1)}\), let \(\Omega(b^{(t-1)})\) be its pixel support and partition it into a 3×3 uniform grid with cell indices \((i,j)\in\{1,2,3\}^2\). For each cell, collect motion vectors whose reference-block centers fall inside, and compute the average:
\[
\colshrink{
\bar d_{ij}=(\bar u_{ij},\bar v_{ij})
=\frac{1}{|\mathcal{V}_{ij}|}\sum\nolimits_{(u,v)\in\mathcal{V}_{ij}}(u,v)\quad\text{(pixels)}.}
\]
\noindent Let \(\mathcal{S}=\{(i,j)\mid |\mathcal{V}_{ij}|>0\}\). We compute the mean translation over non-empty cells
\[
\colshrink{
\begin{aligned}
\bar d=\frac{1}{|\mathcal{S}|}\sum_{(i,j)\in\mathcal{S}}\bar d_{ij},
\quad
\sigma_{\mathrm{tr}}^{2}=\frac{1}{|\mathcal{S}|}\sum_{(i,j)\in\mathcal{S}}\bigl\|\bar d_{ij}-\bar d\bigr\|_2^{2}.
\end{aligned}}
\]
Cells with \(|\mathcal{V}_{ij}|=0\) are skipped when computing \(\bar d\), \(\sigma_{\mathrm{tr}}\), and statistics of scale ratios (e.g., \(\mu_r, \sigma_r\)).
We also use a small \(\varepsilon>0\) to avoid division by zero in \(r_{ij}\).
If \(|\mathcal{S}|=0\), declare propagation failure.

\noindent\textbf{Translation test.}
If \(\sigma_{\text{tr}}\le \tau_{\text{tr}}\), apply a pure translation to the box center:
\[
\colshrink{
\begin{aligned}
x_c^{(t)} &= x_c^{(t-1)} + \bar u / W, &
y_c^{(t)} &= y_c^{(t-1)} + \bar v / H,\\
w^{(t)}   &= w^{(t-1)},                 &
h^{(t)}   &= h^{(t-1)}.
\end{aligned}}
\]

\noindent\textbf{Uniform-scale test.}
If the translation test fails, compute cell centers \(g_{ij}\) relative to the box center in pixels, and their shifted versions \(g'_{ij}=g_{ij}+\bar d_{ij}\).
With a small \(\varepsilon>0\), define the per-cell \emph{scale ratios} and statistics as:
\[
\colshrink{
\begin{aligned}
r_{ij}=\frac{\lVert g'_{ij}\rVert_2}{\lVert g_{ij}\rVert_2+\varepsilon},\qquad
\mu_r=\frac{1}{|\mathcal{S}|}\sum_{(i,j)\in\mathcal{S}} r_{ij},\qquad
\sigma_r^2=\frac{1}{|\mathcal{S}|}\sum_{(i,j)\in\mathcal{S}}\bigl(r_{ij}-\mu_r\bigr)^2.
\end{aligned}}
\]
If \(\sigma_r\le \tau_{\text{sc}}\), update scale and center by the average translation:
\[
\colshrink{
\begin{aligned}
w^{(t)}   &= \mu_r\, w^{(t-1)}, &\qquad
h^{(t)}   &= \mu_r\, h^{(t-1)},\\
x_c^{(t)} &= x_c^{(t-1)} + \bar u / W, &\qquad
y_c^{(t)} &= y_c^{(t-1)} + \bar v / H.
\end{aligned}}
\]

\noindent\textbf{Propagation failure and fallback.}
If neither test is accepted, declare propagation failure at time \(t\) and call OWLv2. Even if a test passes, apply an area-growth check to prevent explosion. Let
\[
\colshrink{
\begin{aligned}
A(b) &= WH\,wh,\\
\operatorname{fallback}_n(t) &= \bigl[\,A\!\left(b_n^{(t)}\right) > 2A_n^\star \ \text{and}\ (t - t_n^\star) \le 10\,\bigr],\\
\text{on fallback (box }n\text{)} &= \bigl(t_n^\star \leftarrow t,\ \ A_n^\star \leftarrow A\!\left(b_n^{(t)}\right)\bigr).
\end{aligned}}
\]
These thresholds were set conservatively based on common motion heuristics to prevent rapid, unrealistic bounding box growth within a short time window, and were not tuned on the target dataset.

\noindent\textbf{Single-class switch.}
After any OWLv2 call, if and only if there is exactly one detection with score \(s \ge \tau_{\text{cls}}\), set \(S_t=1\) and set \(c^\dagger\) to that class for subsequent prompts; otherwise set \(S_t=0\).
During single-class mode, if OWLv2 fallback occurs and returns empty or scores below \(\tau_{\text{cls}}\) for \(M\) consecutive frames, we reset \(S_t=0\).

\noindent\textbf{Clipping and validity.}
After any update, clip \(\Gamma^{-1}(b;W,H)\) to the image bounds, drop boxes with invalid extent or near-zero area, and re-normalize by \(\Gamma\) back to YOLO coordinates.
\end{methodmath}

\medskip\noindent\textbf{Components for ablation.}
We ablate two toggles—(i) \emph{3×3 Grid} MV aggregation and (ii) the \emph{area-growth} consistency check—while keeping the single-class mode on unless stated otherwise; see Sec.~\ref{sec:experiments} for the concrete configurations.
 
\section{Experiments}
\label{sec:experiments}

We evaluate on the ILSVRC2015-VID validation set~\cite{russakovsky2015imagenet} with its 30 categories. 
All open-vocabulary methods use the same list of class prompts (OWLv2~\cite{minderer2023scaling}, YOLO-world~\cite{cheng2024yolo}, and our MVP). Supervised YOLOv12~\cite{tian2025yolov12} models use their predefined label sets and do not use text prompts. 
Our approach is training-free and does not fine-tune on this dataset. 
We report mAP at four IoU settings: 0.2, 0.3, 0.5, and [0.5:0.95]~\cite{lin2014microsoft}, together with throughput in frames per second (FPS) measured on RTX~3090 when available.

\medskip\noindent\textbf{Baselines.}
Framewise OWLv2~\cite{minderer2023scaling} (Large and Base) as open-vocabulary detectors; 
YOLO-world as a zero-shot detector; 
the \emph{YOLOv12 family} (s/m/l/x) as supervised references; 
and classical object trackers (MOSSE, KCF, CSRT) used as propagation modules between sparse keyframes.

\medskip\noindent\textbf{Main results.}
Table~\ref{tab:main} compares accuracy, speed, and capability. 
Framewise OWLv2-Large achieves the strongest accuracy. 
Our method (MVP) is close at loose IoUs while reducing detector invocations, and reaches 10.3\,FPS on RTX~3090. 
Within the YOLOv12 family, accuracy and speed scale with model size from s to x as expected. 
OWLv2-Base in this run covers fewer images than other rows and is therefore not strictly comparable; we keep it only as a sanity check.

\begin{table*}[t]
\centering
\caption{Comparison on ILSVRC2015-VID (val). We report four mAPs and FPS (RTX~3090). }
\label{tab:main}
\resizebox{\linewidth}{!}{
\begin{tabular}{lccccccccc}
\toprule
Method & mAP@0.2 ($\uparrow$) & mAP@0.3 ($\uparrow$) & mAP@0.5 ($\uparrow$) & mAP@[0.5:0.95] ($\uparrow$) & FPS ($\uparrow$) & Category space & Training mode \\
\midrule
YOLOv12s \cite{tian2025yolov12} & 0.556 & 0.551 & 0.521 & 0.379 & 41 & Closed-set  & Supervised  \\
YOLOv12m \cite{tian2025yolov12} & 0.597 & 0.594 & 0.569 & 0.423 & 39 & Closed-set  & Supervised  \\
YOLOv12l \cite{tian2025yolov12} & 0.604 & 0.599 & 0.570 & 0.422 & 37 & Closed-set  & Supervised  \\
YOLOv12x \cite{tian2025yolov12} & 0.651 & 0.648 & 0.631 & 0.485 & 36 & Closed-set  & Supervised  \\
YOLO\text{-}world \cite{cheng2024yolo} & 0.323 & 0.322 & 0.316 & 0.247 & 40 & Open-vocab & Zero-shot \\
OWLv2-Base \cite{minderer2023scaling} & 0.698 & 0.694 & 0.676 & 0.520 & 1.5  & Open-vocab & Zero-shot \\
OWLv2-Large \cite{minderer2023scaling} & 0.784 & 0.780 & 0.760 & 0.552 & 0.8  & Open-vocab & Zero-shot \\
\textbf{MVP (ours)}     & \textbf{0.747} & \textbf{0.721} & \textbf{0.609} & \textbf{0.316} & \textbf{10.3} & \textbf{Open-vocab} & \textbf{Zero-shot} \\
\bottomrule
\end{tabular}}
\vspace{-2pt}
\begin{minipage}{\textwidth}\footnotesize
\textbf{Notes.} 
($\uparrow$) higher is better. FPS values are measured on RTX~3090 under our unified protocol.  
\textit{Closed-set} refers to conventional detectors trained and evaluated on a fixed set of categories (e.g., the 30 classes of ILSVRC2015-VID).  
\textit{Supervised} means these models require category-specific annotated training data.  
In contrast, \textit{Open-vocab} denotes the ability to detect arbitrary categories specified by free-form text prompts, and \textit{Zero-shot} means such generalization is achieved without any task-specific training.  
Together, open-vocabulary and zero-shot capability represent a fundamental leap beyond closed-set supervised detection, enabling recognition of unseen objects without costly manual annotation. 
\end{minipage}
\end{table*}

\medskip\noindent\textbf{Fixed keyframe interval.}
We also evaluate OWLv2-Large under a fixed keyframe interval \(K\) (Table~\ref{tab:intervals}). 
At each keyframe we run the detector; on intermediate frames we simply \emph{reuse} the bounding boxes predicted at the most recent keyframe with positions held fixed (``frozen boxes'', no motion compensation).
As \(K\) grows, this frozen-box baseline trades accuracy—especially at tight IoUs—for higher FPS.
By contrast, MVP propagates boxes using motion cues, preserving coarse localization while avoiding the sharp drop.

\begin{table}[t]
\centering
\caption{OWLv2-Large at fixed intervals (\(K\)) versus our MVP, with FPS on RTX~3090.}
\label{tab:intervals}
\resizebox{\linewidth}{!}{
\begin{tabular}{lccccc}
\toprule
Strategy & mAP@0.2 & mAP@0.3 & mAP@0.5 & mAP@[0.5:0.95] & FPS \\
\midrule
Every 5 frames   & 0.604 & 0.602 & 0.590 & 0.428 & 3.2 \\
Every 10 frames  & 0.544 & 0.543 & 0.531 & 0.371 & 5.4 \\
Every 30 frames  & 0.461 & 0.458 & 0.435 & 0.275 & 11.1 \\
Every 50 frames  & 0.416 & 0.411 & 0.379 & 0.226 & 11.3 \\
\textbf{MVP (ours)} & \textbf{0.747} & \textbf{0.721} & \textbf{0.609} & \textbf{0.316} & \textbf{10.3} \\
\bottomrule
\end{tabular}}
\end{table}

\medskip\noindent\textbf{Propagation module comparison.}
Under the same keyframe schedule, replacing MVP with classical trackers degrades mAP at all IoUs and lowers throughput.

\begin{table}[t]
\centering
\caption{Propagation with classical trackers versus MVP, with FPS on RTX~3090.}
\label{tab:trackers}
\resizebox{\linewidth}{!}{
\begin{tabular}{lccccc}
\toprule
Method & mAP@0.2 & mAP@0.3 & mAP@0.5 & mAP@[0.5:0.95] & FPS \\
\midrule
MOSSE~\cite{bolme2010visual} & 0.629 & 0.574 & 0.426 & 0.205 & 4.8 \\
KCF~\cite{henriques2014high} & 0.684 & 0.646 & 0.513 & 0.248 & 4.8 \\
CSRT~\cite{lukezic2017discriminative} & 0.620 & 0.558 & 0.395 & 0.180 & 6.6 \\
\textbf{MVP (ours)} & \textbf{0.747} & \textbf{0.721} & \textbf{0.609} & \textbf{0.316} & \textbf{10.3} \\
\bottomrule
\end{tabular}}
\end{table}

\medskip\noindent\textbf{Ablation.}
We ablate two toggles—(i) \emph{3×3 Grid MV aggregation} and (ii) the \emph{area-growth check}—while keeping the single-class mode \emph{on} unless otherwise noted. 
Table~\ref{tab:ablation_merged} merges configurations (\cmark/\xmark) and results; directory names from our logs are shown for reproducibility.

\begin{table}[t]
\centering
\caption{Ablation of MVP components on ILSVRC2015-VID (val). A \cmark means the component is enabled. FPS is wall-clock on RTX~3090 including decoding, MV access, preprocessing, and non-maximum suppression (NMS).}
\label{tab:ablation_merged}
\setlength{\tabcolsep}{4pt}
\resizebox{\linewidth}{!}{
\begin{tabular}{@{}lccccccc c@{}}
\toprule
\textbf{Setting} & \textbf{3×3 Grid MV} & \textbf{Area-growth} & \textbf{Single-class} 
& \textbf{mAP@0.2} & \textbf{mAP@0.3} & \textbf{mAP@0.5} & \textbf{mAP@[0.5:0.95]} & \textbf{FPS} \\
\midrule
\textit{w/o Single-class check} & \cmark & \cmark & \xmark & 0.747 & 0.721 & 0.609 & 0.316 & 10.1 \\
\textit{w/o Area-growth check}          & \cmark & \xmark & \cmark & 0.736 & 0.710 & 0.601 & 0.315 & 10.3 \\
\textit{w/o 3×3 Grid MV} & \xmark & \cmark & \cmark & 0.524 & 0.460 & 0.309 & 0.136 & 25.0 \\
\textbf{MVP (full)} & \cmark & \cmark & \cmark & \textbf{0.747} & \textbf{0.721} & \textbf{0.609} & \textbf{0.316} & \textbf{10.3} \\
\bottomrule
\end{tabular}}
\vspace{-2pt}
\begin{minipage}{\columnwidth}\footnotesize
\textbf{Notes.} 
Disabling the single-class switch changed FPS by $<\!2\%$ and left mAP unchanged in our setup (30-class prompts with text-embedding caching); we include the row here for completeness. 
\end{minipage}
\end{table}

\medskip\noindent\textbf{Takeaways.}
(1) Framewise OWLv2-Large is the accuracy upper bound.  
(2) MVP preserves most coarse detections at 0.2 and 0.3 with an expected drop at tight IoUs.  
(3) Under the same schedule, MVP is stronger and faster than tracker-based propagation.  
(4) Fixed-interval sampling shows a clear accuracy versus interval tradeoff; MVP avoids the large loss seen with very sparse detection while keeping double-digit FPS.  
(5) Within the YOLOv12 family, accuracy and FPS scale with model size; these supervised baselines require labeled training, whereas MVP is training-free and open-vocabulary.

\section{Discussion and Conclusion}

Our results indicate that a training-free, zero-shot video detector is practical when paired with compressed-domain motion. Invoking OWLv2~\cite{minderer2023scaling} at fixed keyframe intervals and propagating boxes with codec motion vectors retains much of the framewise detector coarse localization while reducing redundant inference. 

\medskip\noindent\textbf{Discussion.}
Prompted OWLv2 provides broad category coverage without dataset-specific training, detecting classes absent from conventional supervised label spaces. Under matched keyframe intervals, MVP remains close to framewise OWLv2-Large at loose IoUs (0.2/0.3) but, as expected for propagation, degrades at tight IoUs; thus it is suitable when coarse localization suffices or compute is constrained. Under the same schedule, MVP outperforms classical tracker propagation (MOSSE, KCF, CSRT) in mAP across IoUs; tracker rows serve as sanity checks, not primary baselines. Supervised baselines (YOLOv12s/m/l on ImageNet-VID) may surpass MVP at IoU~0.5, but require labels and lack open-vocabulary capability.

\medskip\noindent\textbf{Limitations.}
The approach depends on the quality of codec motion vectors: heavy compression, very low resolution, strong camera motion, or complex deformation can introduce noise and trigger more frequent re-detections. High-precision localization of small or fast-deforming objects, and scenes with large parallax, remains challenging when relying only on translation and uniform scaling. Prompt design is also important, since overly broad or poorly calibrated prompts may increase false positives or cause unnecessary fallbacks. 

\medskip\noindent\textbf{Future work.}
We plan to explore content-aware keyframe scheduling with lightweight compressed-domain signals; stronger propagation via multi-grid models, residual cues, or confidence-aware fusion to improve tight-IoU accuracy while remaining training-free; simple re-identification to maintain long-range consistency after occlusion; and prompt calibration or automatic prompt expansion to stabilize open-vocabulary coverage over long clips.

\medskip\noindent\textbf{Conclusion.}
We presented a training-free, open-vocabulary video detector that couples OWLv2 with motion-aware propagation at fixed keyframe intervals. On ILSVRC2015-VID, the method preserves much of the framewise accuracy at loose IoUs, consistently surpasses tracker-based propagation under the same schedule, and requires no additional labels or fine-tuning—making it a practical building block for scalable video understanding when both flexibility and efficiency matter.

%
%
%
%

\clearpage

\bibliographystyle{IEEEbib}
\bibliography{strings,refs}

\end{document}